# ROBUST STATISTICAL APPROACH FOR EXTRACTION OF MOVING HUMAN SILHOUETTES FROM VIDEOS

Oinam Binarani Devi, Nissi S Paul, Y Jayanta Singh

Computer Science & Engineering and Information Technology, Don Bosco College of Engineering and Technology, Assam Don Bosco University, Guwahati, Assam,

**ABSTRACT**

*Human pose estimation is one of the key problems in computer visionthat has been studied in the recent years. The significance of human pose estimation is in the higher level tasks of ―understanding human actions applications such as recognition of anomalous actions present in videos and many other related applications. The human poses can be estimated by extracting silhouettes of humans as silhouettes are robust to variations and it gives the shape information of the human body. Some common challenges include illumination changes, variation in environments, and variation in human appearances. Thus there is a need for a robust method for human pose estimation. This paper presents a study and analysis of approaches existing for silhouette extraction and proposes a robust technique for extracting human silhouettes in video sequences. Gaussian Mixture Model (GMM) A statistical approach is combined with HSV (Hue, Saturation and Value) color space model for a robust background model that is used for background subtraction to produce foreground blobs, called human silhouettes. Morphological operations are then performed on foreground blobs from background subtraction. The silhouettes obtained from this work can be used in further tasks associated with human action interpretation and activity processes like human action classification, human pose estimation and action recognition or action interpretation.*

**KEYWORDS**

*Background subtraction, background modelling, human silhouette extraction, dilation, erosion*

## 1. INTRODUCTION

Analysis of human actions for detecting and recognizing of anomalous actions present in videos is an active research area in computer vision. The detection of anomalous actions will benefit in the application of healthcare organization for old age homes and clinical diagnosis of patients. Foreground segmentation is one of the fundamental tasks in a computer vision system for all applications like video surveillance, understanding human walking motion and other related applications. The video sequences containing humans in motion, segmentation plays a crucial role in analysing human motion tracking and action recognition as it separates the object of interest before extracting information such as temporal data, velocity, poses and so on. A reliable segmentation method is required for robust human body segmentation from a video sequence captured by a static and fixed camera in indoor as well as outdoor environments. The segmented human silhouettes are used in analysis of shape and posture of human actions and many high-level vision processing tasks relating to human activity analysis, activity modelling, mobility assessment and abnormal event detection. Human silhouette extractions are possible on both simple-static and complex- dynamic background. To obtain the human silhouettes, commonly background subtraction method is used. There are many variations of background subtraction methods available in the literature to handle illumination changes, shadow removal, occlusions removal. The background subtraction method used has to be a novel and adaptive one that extracts foreground objects accurately.

This paper combines techniques of background subtraction using the features based on HSV (Hue Saturation and Value) and post processing operations to get robust silhouettes. Silhouettes help in further processes like classification, tracking and action interpretation.

The proposed method consists of three steps: (1) pre-processing, (2) segmentation and (3) post-processing.

(1) Pre-processing is a method of normalizing. Converting the RGB (Red, Green and Blue) into normalized RGB removes the effect of any intensity variations. It helps in comparing frames taken under variations of illumination.
(2) Segmentation is the process of partitioning the part of an image where the human silhouette is extracted. The background subtraction method is used.
(3) Post-processing is a method that includes morphological operations for refining and enhancing the obtained results of segmentation.

There are three varying approaches as given below and in all these approaches the RGB are converted to HSV and only the third layer of HSV i.e. V (Value) layer is considered. The Value layer is considered because it is more stable than Hue and Saturation (which is explaining in Section 3). The following are the approaches that have been experimented in this paper:

**Approach I:** A background model is generated using HSV colour space model on the initial frame which does not contain human in motion and the incoming consisting of humans in motion are differed from the initial frame which is modelled to extract the silhouette of the moving human.

**Approach II:** In the second approach background model is generated using a Simple Gaussian method where, mean and variance of the pixels of the background in HSV colour space are computed. The incoming consisting of humans in motion are differed from the frame which is modelled. The Gaussian model [1] for each pixel is the Gaussian density as described below:

$$P(x) = \frac{1}{\sqrt{2\pi\sigma}} \exp\left[-\frac{1}{2}\left(\frac{x-\mu}{\sigma}\right)^2\right] \quad \text{------- (1)}$$

where $P(x)$ is the density of the pixels, $\mu$ is the mean value of pixels and $\sigma$ is the variance.

**Approach III:** The third approach is a variation of Stauffer and Grimson's method of Gaussian Mixture [2]. In this paper, the proposed method uses background modelling of Gaussian Mixture Model on HSV color space of the frame. The HSV color space model helps in handling illumination changes maintain stability of the image. The incoming consisting of humans in motion are differed from the frame which is modelled. In Gaussian mixture model, each pixel in each frame is distributed by K mixture in Gaussian Model [1] and the probability of the pixel $x_t$ at time $t$ is obtained as:

$$P(x_t) = \sum_{i=1}^{K} w_i \eta(x_t, \theta_i) \quad \text{-------- (2)}$$

where, $w_i$ weight parameter of $K$ Gaussian factor and $\eta(x_t, \theta_i)$ is the normal distribution of K.

The normal distribution is given by:

$$\eta(x_t, \theta_i) = \eta(x, \mu_k, \Sigma_k) \frac{1}{(2\pi)^{\frac{D}{2}} |\Sigma_k|^{\frac{1}{2}}} \exp(Z) \quad \text{-------- (3)}$$

Where, $\mu_k$ is the mean and $\Sigma_k$ covariance of $K$ factor and Z is found out by:

$$Z = \left[ -\frac{1}{2}(x - \mu_k)^T \Sigma_k^{-1} (x - \mu_k) \right] \quad \text{-------- (4)}$$

$K$ is the number of distribution estimated according to the $w_i$ and $T$ is the threshold and 0.2 is used as the threshold value to model the background.

The rest of the paper is organized as follows, Section 2 presents the related works done, followed by the design and implementation of the system, and its descriptions. Section 3 gives
the results obtained and Section 4 draws the conclusion with Section 5 giving the future works th at can be extended and Section 6 gives the references.

## 2. RELATED WORKS

In the work of Chen et al. [3] features are extracted in colour space accumulating the feature information over a short period of time and fuse high-level with low-level information by building a time varying background model. In their work to separate silhouettes of moving objects from a human body silhouette a fuzzy logic inference system is developed handling the challenges of shadows and capable of operating in real-world, unconstrained environments with complex and dynamic backgrounds.

In the paper of Scheer et al.[4] , silhouettes are extracted in a YUV (Y is luminance, U and V are chrominance or colour components) colour space which is able to detect shadow but this algorithm does not work well. When the target object is not in motion or when there is presence of other moving objects in the scene.

In the work of Setiawan et. al. [5] Gaussian Mixture Model (GMM) for foreground segmentation is used along with an IHLS (Improved Hue, Light and Saturation) colour space model which can differentiate shadow region from objects. IHLS colour space model serves as the fundamental description for image as it has an advantage over Red Green Blue (RGB) colour space to recognize shadow region from the object by utilizing luminance and saturation-weighted hue information directly without any calculation of chrominance and luminance.

In Seki et al [6] work, method of distribution of image vector is used as feature from the colour change in the observed background. This vector subtracts from the background based on the Mahalanobis distance between illumination intensity, the reflection index of objects and pixel values with the assumption that there is no change in the distribution of illumination intensity in a small region over some time.

In this paper a robust background subtraction method is proposed to obtain the moving human silhouettes. Accurate human silhouettes are needed to extract the features of human body configuration as done in the work of Dedeoglu et. al. [7] . In most of the existing research work on moving object detection the methods done by Collin et. al. [8], Wang and David [9] to locate people and their parts (head, hands, feet and torso) as in the work of Wang et. al.[10] background subtraction is used which can be known from the reviews done by Alan [11] and Piccardi [12]. In Haritauglu et al. [13] , approaches for detecting and tracking moving objects in videos from a static camera background subtraction is used as well as it serves as one of the central tasks. For a robust background subtraction, the RGB (Red, Green and Blue) color frames are converted to HSV (Hue, Saturation and Value) color space in the works of Cucchiara et. al. [14] and Zhao et. al. [15] .The background subtraction method has been used to aid in human behaviour analysis and action interpretation and in recent methods background subtraction is used to obtain human silhouettes.

## 3. DESIGN AND IMPLEMENTATION

Public datasets of Weizmann [13], Visor [14] and UT-Interaction [15] dataset consisting of single persons walking are used with the videos captured by static camera and fixed background environment. Our design of the proposed work for the extraction of moving silhouettes is illustrated in Figure 1.

### 3.2. Design of the human silhouette extraction system

The design part consists of taking in the sequences of frames from the video input from which the RGB will be normalized and converted to HSV. The processes are explained below:

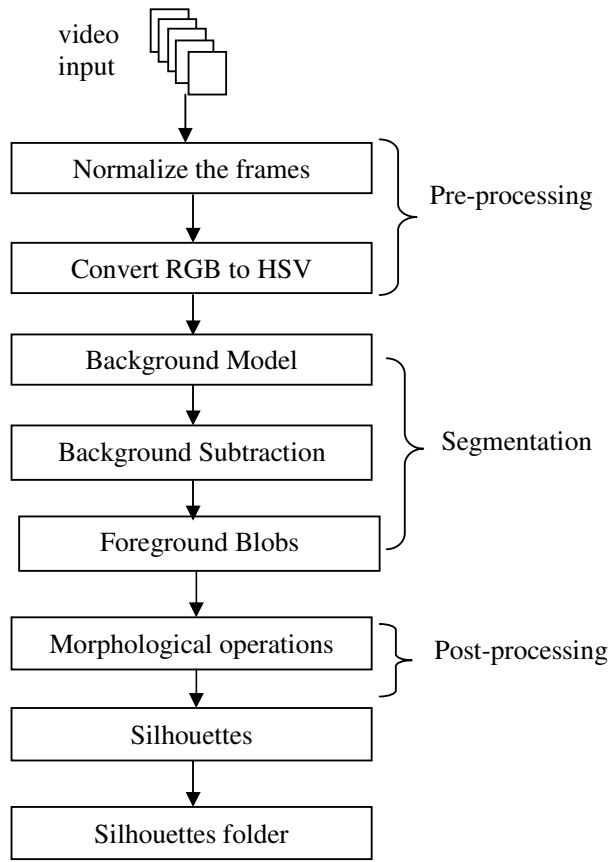

Fig. 1. Proposed method

International Journal on Information Theory (IJIT), Vol.3, No.3, July 2014International Journal on Information Theory (IJIT), Vol.3, No.3, July 2014

### 3.2.1. Generation of background model

Most of the videos captured are in RGB (Red, Green and Blue) colour model. In the proposed work, the input video in RGB colour space model is converted to HSV (Hue, Saturation and Value) colour space, consisting of three layers hue, saturation and value. In RGB colour space the three color components are correlated and is found to be disadvantageous as it is difficult to separate the colour information from the brightness. When one is affected by illumination changes the remaining layers will also be affected leading to instability. In order to overcome the instability caused by RGB, HSV colour space is used to differentiate brightness from chromacity.

In Figure 2 below, The RGB to HSV converted frame is decomposed into their respective Hue, Saturation and Value layers and it is clearly noticeable that Hue is not affected by illumination and value is more stable than Hue and Saturation.

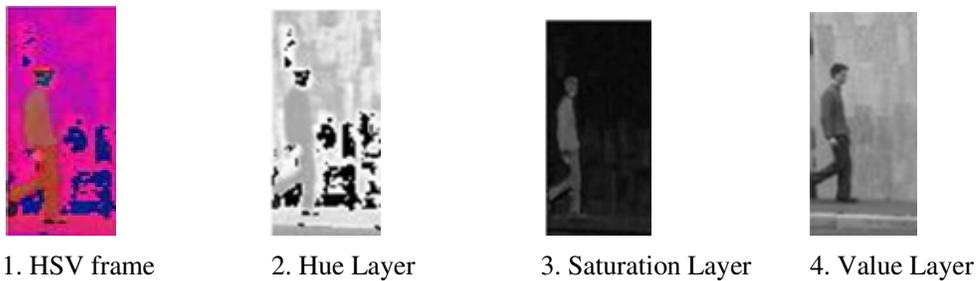

1. HSV frame         2. Hue Layer         3. Saturation Layer         4. Value Layer

Fig. 2. Decomposition of Hue Saturation Value layer of an Red Green Blue frame

The following is the histogram distribution of the HSV layers which display the density of the pixels distributed in each of the layer.

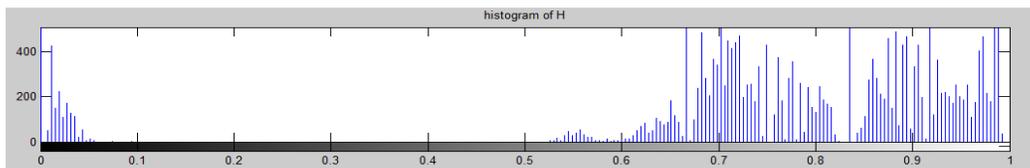

1. Histogram of Hue Layer

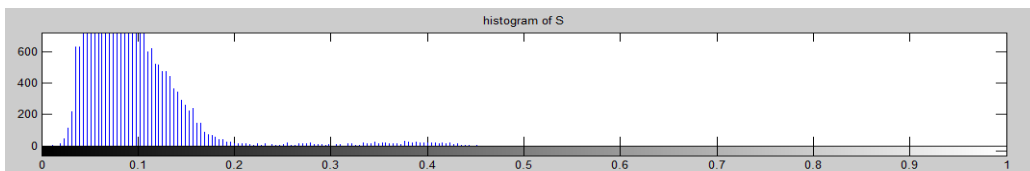

2. Histogram of Saturation Layer

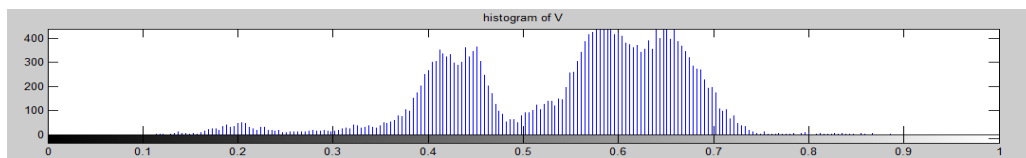

3. Histogram of Value Layer

Fig. 3. Histogram distribution of 1. Hue, 2. Saturation and 3. Value Layer (top to bottom).

59

In all the above histogram distribution figures the horizontal axis is the range of pixels and the vertical axis is the number of times the pixels are occurring in the image. In Fig 3,

1. Histogram of Hue layer shows the distribution of the color pixels towards the whitish part.
2. Histogram of Saturation layer shows the purity of the colors and the color pixels are densely distributed to the darker part which is why Fig 2, Saturation Layer is dark.
3. Histogram of Value layer gives the uniform distribution of the color pixels.

The third Value layer consists of the maximum information of the three Red, Green or Blue. It can be shown as equation (7) .With the help of this HSV colour space and the Value layer model these approaches of the background modelling are experimented.

### 2.2.2. Segmentation

Segmentation is the process of separating the foreground pixels from the background pixels thereby extracting the object of interest. Once the background model is generated, foreground blobs are obtained by background subtraction method. The background subtraction results in the segmentation of the foreground pixels from the background and the human shape foreground blobs are obtained.

### 2.2.3. Morphological operations

Finally, morphological operations- dilation and erosion are applied to get clean silhouettes of the human body shape. It is the post-processing technique that helps in eliminating the noisy pixels. Dilation operation helps in adding pixels to the boundary of the foreground blobs and erosion operation shrinks the foreground blobs.

## 3.3. Implementation Steps for the human silhouette extraction system

The following are the steps in details that are implemented in the human silhouette extraction system.

### 3.3.1. Normalization of the RGB colour video [19]

Normalization is done to remove the effect of any intensity variations which helps in comparing taken under variations of illumination and the range of the colors will be changed from [0,255] to [0,1].

$$r` = r/255; g` = g/255; b` = b/255;$$
$$cmax = \max(r`,g`,b`);$$
$$cmin = \min(r`,g`,b`);$$
$$\Delta = cmax - cmin$$

Where,

r, g and b represent the Red, Green and Blue color layers,
r`, g` and b` are the normalized Red, Green and Blue.
max (r`,g`,b`) finds out the maximum among r`, g` and b` and similarly min (r`,g`,b`) finds out the minimum among r`, g` and b` which are assigned to the corresponding cmax and cmin.
 $\Delta$ is the difference between cmax and cmin.

### 3.3.2. RGB to HSV [19]

The Hue depicted by h, Saturation depicted by s and Value depicted by v is converted from the respective normalized values of Red (r`), Green (g`) and Blue (b`).

1. Hue calculation

$$h = \begin{cases} 60 \times \left(\dfrac{g`-b`}{\Delta}, \mod 6\right), & c\max = r` \\ 60 \times \left(\dfrac{b`-r`}{\Delta} + 2\right), & c\max = g` \\ 60 \times \left(\dfrac{r`-g`}{\Delta} + 4\right), & c\max = b` \end{cases} \quad\text{------ (5)}$$

2. Saturation calculation

$$s = \begin{cases} 0, & \Delta = 0 \\ \dfrac{\Delta}{c\max}, & \Delta \neq 0 \end{cases} \quad\text{------ (6)}$$

3. Value calculation

$$v = c\max \quad\text{------- (7)}$$

### 3.3.3. Background Modelling, Background Subtraction, Foreground Blobs

The background model will be generated from the initial frame or from a frame which is an empty scene. This background model is used for background subtraction with human presence in the scene. The background subtraction results in obtaining foreground blobs which serves the required silhouettes.

### 3.3.4. Morphological Operations

Morphological operations are applied onto the resulted foreground blobs using dilation and erosion methods to remove the noisy pixels and obtain accurate human silhouettes.

## 4. RESULTS

The above three approaches were experimented on the video datasets of (i)Weizmann, (ii)Visor and (iii) UT-Interaction of a single person in walking action. The silhouettes obtained are represented in figure 4 where the first, second and third columns show the silhouette of Weizmann, Visor and UT-interaction video respectively. The results of first row are obtained by the Approach I of using simple background modelling and background subtraction in HSV color space and considering only the Value layer of HSV. The second row consists of the Approach II where a Simple Gaussian background model is generated and background subtraction is performed in the same manner of using HSV color space and taking only the Value layer of HSV. The third row shows the result of the proposed Approach III in which the video sequences

61

International Journal on Information Theory (IJIT), Vol.3, No.3, July 2014

are using Gaussian Mixture Model background model along with incoming subtracted from the background model generated on the HSV converted using the third layer.

| Background Subtraction Methods | Chosen Data Sets and sample data | | |
|---|---|---|---|
| | Weizmann Walk dataset[16] | Visor Dataset [17] | UT-Interaction Dataset[18] |
| Approach I: Frame differencing | 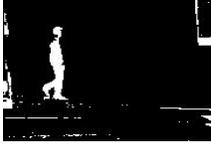 | 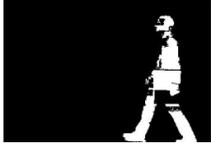 | 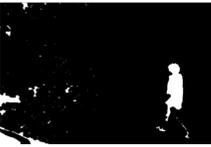 |
| Approach II: Simple Gaussian Method | 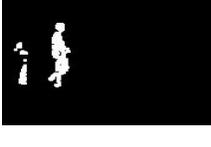 | 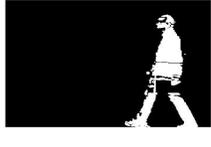 | 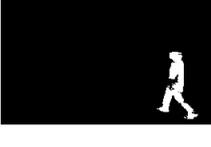 |
| Approach III: Gaussian Mixture Model | 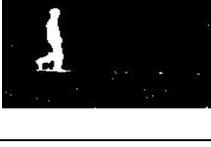 | 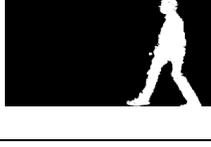 | 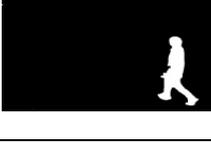 |

Figure4: The sample of the silhouettes obtained using each approaches

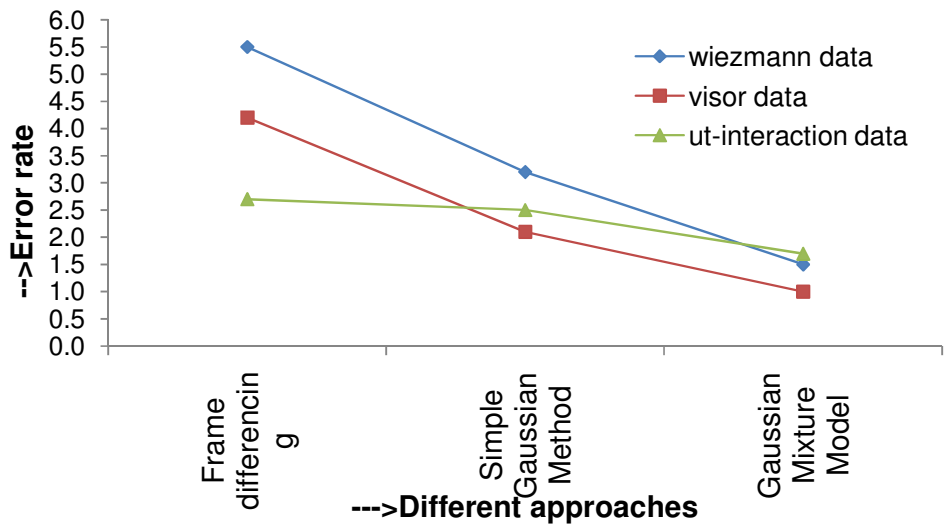

Figure5: The error percentage in each of each approaches (Statistical data only)

This study compares the percentage of error generated by each of the approaches with the chosen three datasets and shown in figure5

a) Approach I: Frame differencing
This approach is applied to three chosen dataset. The error rate generated by this approach on Wiezmann dataset is 5.5%, Visor dataset is 4.2% and Ut-interaction dataset is 2.7%. These error rate is compare to the next two approaches.

b) Approach II: Simple Gaussian method
The error rate generated by this approach on Wiezmann dataset is 3.2%, Visor dataset is 2.1% and Ut-interaction dataset is 2.5%. By using Wiezmann dataset, this study could minimised the error rate half of those error generated in first approach. However error rate is varying from one dataset to another.

c) Approach III: Gaussian Mixture Model
The error rate generated by this approach on Wiezmann dataset is 1.5%, Visor dataset is 1% and Ut-interaction dataset is 1.7%. Here also the error rate are varied from one dataset to another. However this approach i.e. Gaussian Mixture Model has lesser error rate in all the dataset chosen and gives a better performance over the above two approaches.

## 5. CONCLUSIONS

The HSV colour model and the proposed robust background subtraction technique presented in this paper produce human silhouettes that adapt to indoor and outdoor environments with various illumination changes. A statistical approach of background modelling - Gaussian Mixture Models is used. It is seen that the results obtained by approach III, i.e background modelling of Gaussian Mixture Model on HSV color space of the frame gives a better outcome than the other two approaches. The comparison  The silhouettes so obtained can be used for human action classification, human pose estimation and action recognition or action interpretation.

## 6. FUTURE WORK

This work can be further extended by considering the area and centroid of the silhouettes for further processes of analysis like classification, pose estimation and action interpretation. These features will help in serving the properties of the silhouettes in determining the changes occurring in the shapes. The change in shape will help in analysing and interpreting human action.

## 7. REFERENCES

1. Bouwmans T., El Baf, Vachon B. (2008), Background Modeling using Mixture of   Gaussians for Foreground Detection – A Survey, Recent Patents on Computer Science 1,3   219-237.
2. Stauffer C. and Grimson W.E.L., "Learning Patterns of Activity Using Real-Time   Tracking," IEEE Trans. Pattern Analysis and Machine Intelligence, vol. 22, no. 8, pp. 747- 757.
3. Chen X., Zhihai He, Derek A., James K., and Skubic M.(2006), Adaptive Silhouette Extraction  and Human Tracking in Dynamic Environments1, IEEE Transactions on   Circuits and  System for Video Technology.
4. Schreer O., Feldmann I., Golz U. and Kauff P. (2002), Fast and Robust Shadow Detection  in Videoconference Applications.
5. Setiawan N. A., Seok-Ju H., Jang-Woon K., and Chil-Woo L.(2006), Gaussian Mixture  Model in Improved HLS Color Space for Human Silhouette Extraction.

6. Seki M., Fujiwara H., and Sumi K. (2000), "A Robust Background Subtraction Method for Changing Background," Proc. IEEE Workshop Applications of Computer Vision, pp. 207-213.
7. Dedeoglu Y., Toreyin B., Gudukbay U., and Cetin A. (2006), Silhouette-based method for object classification an human action recognition in video, Computer Vision in Human-Computer Interaction, pp. 64-77.
8. Collins R., Gross R., and Shi J.( 2002), Silhouette-based human identification from body shape and gait, AFG.
9. Wang L., David S. (2007), Recognizing Human Activities from Silhouettes: Motion Subspace and Factorial Discriminative Graphical Model.
10. Wang L., Tan T., Ning H. and Hu W. (2003), "Silhouette Analysis-Based Gait Recognition for Human Identification", IEEE Trans. Pattern Analysis and Machine Intelligence, pp.1505-1518.
11. Alan M. McIvor (2000), Background Subtraction Techniques, Proc. of Image and Vision Computing.
12. Piccardi M.(2004), Background subtraction techniques: a review, Systems, Man and Cybernetics, IEEE International Conference on (Volume : 4).
13. Haritaoglu I., Harwood D., and Davis L.S. (2004), W4: Real-Time Surveillance of People and Their Activities, IEEE Trans. Pattern Analysis and Machine Intelligence, vol. 22, no. 8, pp. 809-830, Aug.2000.
14. Cucchiara R., Grana C., Piccardi M., Prati A., and Sirotti S.(2001), Improving Shadow Suppression in Moving Object Detection with HSV Color Information, Proc. IEEE Int'l Conf. Intelligent Transportation Systems.
15. Zhao M, Jiajun Bu, Chun Chen (2002), Robust background subtraction in HSV color space, Multimedia systems and applications.
16. Weizmann Dataset. http://www.openvisor.org/
17. Visor Dataset. http://www.openvisor.org/
18. Ut-interaction Dataset. http://cvrc.ece.utexas.edu/SDHA2010/Human_Interaction.html
19. RAPIDTABLES, http://www.rapidtables.com/convert/color/rgb-to-hsv.htm

**Authors**

**O. Binarani Devi** is pursuing Master of Technology from the Department of Computer Science & Engineering and Information Technology, Don Bosco College of Engineering and Technology of Assam Don Bosco University, Assam, India. She has completed B.E (Computer Science Engineering) from PGP College of Engineering and Technology of Anna University, Chennai in 2012.

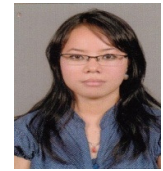

**S. Nissi Paul** is a currently pursuing research in Artificial Intelligence and computer Vision from the Department of Computer Science & Engineering and Information Technology, Don Bosco College of Engineering and Technology of Assam Don Bosco University, Assam, India. She has completed M.Phil (Comp. Sc.) from Bharatidasan University in 2005.

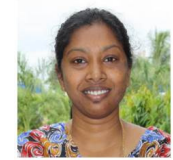

**Y. Jayanta Singh** is working as Associate Prof. and Head of Department of Computer Science & Engineering and Information Technology, Don Bosco College of Engineering and Technology of Assam Don Bosco University, Assam, India. He has received Ph.D.(Comp. Sc. and IT) from Dr Babasaheb Ambedkar Marathwada University, Maharashtra in 2004. He has worked with Swinburne University of Technology (AUS) at (Malaysia campus), Misurata University(North Africa), Skyline University (Dubai), Keane Inc (Canada) etc. His areas of research interest are Real Time Distributed Database, Cloud Computing, Digital Signal processing, Expert Systems etc. He has published several papers in International and National Journal and Conferences. He is presently executing AICTE sponsored Research Project.

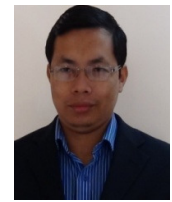